\documentclass[11pt]{article}
\usepackage{geometry}
\usepackage{coling2020}
\usepackage{times}
\usepackage{url}
\usepackage{float}
\usepackage{latexsym}
\usepackage{microtype}
\usepackage{tabularx}
\usepackage{graphicx}

\colingfinalcopy 

\title{Garain at SemEval-2020 Task 12: Sequence based Deep Learning for Categorizing Offensive Language in Social Media}

\author{Avishek Garain \\
  Computer Science and Engineering \\
  Jadavpur University \\
  Kolkata, India \\
  {\tt avishekgarain@gmail.com}}

\date{}

\begin{document}
\blfootnote{\noindent This work is licensed under a Creative Commons Attribution 4.0 International License. License details:\\
http://creativecommons.org/licenses/by/4.0/.}
\blfootnote{
https://sites.google.com/site/offensevalsharedtask/offenseval2019}
\blfootnote{https://sites.google.com/site/offensevalsharedtask/home}
\maketitle
\begin{abstract}
SemEval-2020 Task 12 was OffenseEval: Multilingual Offensive Language Identification in Social Media \cite{zampieri-etal-2020-semeval}. The task was subdivided into multiple languages and datasets were provided for each one. The task was further divided into three sub-tasks: offensive language identification, automatic categorization of offense types, and offense target identification. I have participated in the task-C, that is, offense target identification. For preparing the proposed system, I have made use of Deep Learning networks like LSTMs and frameworks like Keras which combine the bag of words model with automatically generated sequence based features and manually extracted features from the given dataset. My system on training on 25\% of the whole dataset achieves macro averaged f1 score of  47.763\%.
\end{abstract}

\section{Introduction}
\label{intro}
The ever-growing amount of user-generated data on social media platforms be it Facebook, Twitter, blogs or any other electronic medium introduces new challenges in terms of automatic
content moderation, especially regarding hate speech  and  offensive language detection. Not only is hate speech more likely to happen on the Internet,  where anonymity is easily obtained and speakers are psychologically distant from their audience, but its online nature also gives it a far-reaching and determinative impact\cite{shaw2011hate}. User content mostly consists of microposts, where the context of a post can be missing or inferred only from current events. Manual verification of each posting by a human moderator is infeasible due to the high amount of postings created every day. Consequently, automated detection of such attacking postings is the only feasible way to counter this kind of hostility. However, this task is challenging because natural language is fraught with ambiguities, and language in social media is extremely noisy. The classification system that would be prepared for the task, needed to be generalized for various test corpora as well. In this paper I have described the system consisting of a sequential pipeline with text feature extraction and classification as its main components. Firstly, a bag-of-words model is used for encoding the sentences into corresponding integer sequence. Thereafter, vectors are generated from these sequences and fed to a series of BiLSTM layers for training. Then a softmax layer is used for ternary classification into the corresponding offensive language categories.

The rest of the paper has been organized as follows. Section \ref{sec:data} describes the data, on which, the task was performed. The methodology followed is described in Section \ref{sec:methodology}. This is followed by the results and concluding remarks in Section \ref{sec:results} and \ref{sec:conclusion} respectively.
%
%
    %
    %
    %
    %
    %
    %

\section{Data}
\label{sec:data}The dataset that has been used to train and validate the model is an annotation following the tagset proposed in the Offensive Language Identification Dataset (OLID) \cite{OLID} and has been used in OffensEval 2019. This year more accurate and quantified dataset was generated and annotated using a semisupervised approach to increase the quality and quantity of the dataset named the Semi-Supervised Offensive Language Identification Dataset (SOLID)\cite{rosenthal2020large}. The resulting dataset that is the newer version of the  OLID dataset has been used for training and validating purposes. It was collected from Twitter; the data being retrieved the data using the Twitter API by searching for keywords and constructions that are often included in offensive messages. The vast majority of content is not offensive so different strategies have been tried to keep a reasonable number of annotations in the offensive class amounting to around 26\% of the dataset.

The dataset provided consisted of annotations in their original form along with the corresponding labels. Subtask C consisted of the labels \texttt{IND}, \texttt{GRP} and \texttt{OTH}.

\begin{table}[H]
\center
\begin{tabularx}{.49\textwidth}{| l | X |}
\hline
\bf Label & \bf Meaning \\

\hline
IND & Offensive tweet targeting an individual \\
GRP & Offensive tweet targeting a group \\
OTH & Offensive tweet targeting neither group or individual \\
\hline
\end{tabularx}
\caption{Meaning of the labels used in the dataset}
\label{tab1:meaning}
\end{table}
\vspace{-1em}
\noindent The dataset has 188742 instances. Random selection of 50000 instances have been made which are divided into 35000 training data instances and 15000 validation data instances. The test data comprises of 850 instances.

\begin{table}[H]
\center
\begin{tabular}{|c|c|c|r|}
\hline
\bf Label & \bf Train & \bf Validation & \bf Total \\ 
\hline
IND & 28319 & 12136 & 40455 \\
OTH & 2062 & 884 & 2946 \\
GRP & 4619 & 1980 & 6599 \\
\hline
\bf All & 35000 & 15000 & 50000\\
\hline
\end{tabular}
\caption{Distribution of the labels in the dataset}
\label{tab1:dataset-dist}
\end{table}
\vspace{-1em}
\begin{table}[H]
\center
\begin{tabular}{|c|c|}
\hline
\bf Label & \bf Test \\ 
\hline
IND & 580 \\
OTH & 80 \\
GRP & 190 \\
\hline
\bf All & 850\\
\hline
\end{tabular}
\caption{Distribution of the labels in Test dataset}
\label{tab1:dataset-test}
\end{table}
\vspace{-1em}
\section{Methodology}
\label{sec:methodology}
\vspace{-0.5em}
My approach is to preprocess the annotations and then convert the annotations into a sequence of words and convert them into word embeddings. I then run a neural-network based algorithm on the processed tweet. Label based categorical division of data is given in Table~\ref{tab1:dataset-dist}. I have used SenticNet5\cite{Cambria2018SenticNet5D} for finding sentiment values of individual words. The sentiment features play a vital role in context of offensive language as it is during sad state of mind  and hatred towards someone that offensiveness is at its peak. The Sentiment values range from -1 to 1 depicting Negative, Neutral and Positive sentiments. The use of BiLSTM networks in our model might have resulted in better results. The work done by Sepp Hochreiter et al. \cite{10.1162/neco.1997.9.8.1735} led the road-map to newer domain of work by bringing the concept of memory into usage for sequence based learning problems. \\
I first have taken the annotations and sent the raw data through some pre-processing steps, for which I took inspiration from the work on Hate Speech against immigrants in Twitter\cite{garain2019titans}, part of SemEval2019. The steps used here are built as an advancement of this work. \\It consisted of the following steps:
\vspace{-0.5em}
\begin{enumerate}
    \item Replacing emojis and emoticons by their corresponding meanings\cite{garain2019humor}
    \vspace{-0.5em}
    \item Removing mentions
    \vspace{-0.5em}
    \item Replacing words like "XX" or "XXX" with "sexual"
    \vspace{-0.5em}
    \item Contracting whitespace
    \vspace{-0.5em}
    \item Removing URLs
\end{enumerate}

In step 1, for example,\\
",-) " is replaced by "winking happy"\\
":-C" is replaced with "real unhappy"\\
";-(" is replaced with "crying"\\
Similarly I replaced 110 emoticons by their feelings.\\

\noindent The step 3 consists of manually identifying a feature of inclusion of words "XX" or "XXX" in sexual harassment cases and taking advantage of this identified feature. Using this extraction, contributed to increasing accuracy in classification to some extent. They played an important role during the model-training stage.

\begin{figure}[H]
    \centering
    \includegraphics[height=7cm,width=0.5\textwidth]{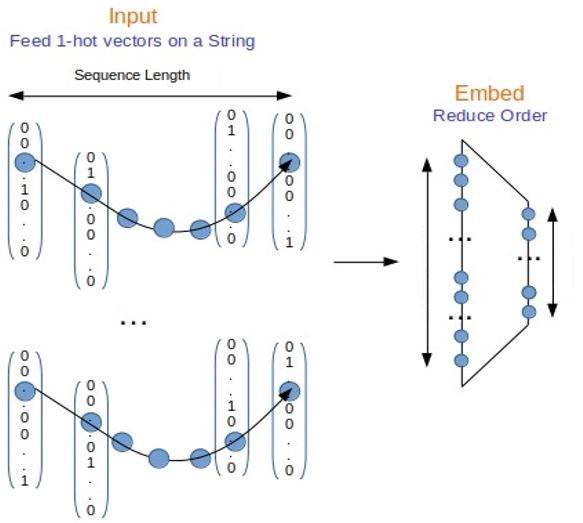}
    \caption{Bag-of-words model}
    \label{fig:bow}
\end{figure}

\noindent The pre-processed annotations are treated as a sequence of words with interdependence among various words contributing to its meaning. I then take a Bag-of-words model approach as well as use pre-trained GloVe vectors. For the Bag-of-words approach I convert the annotations into one-hot vectors as shown in Fig. \ref{fig:bow}. The Bag-of-words approach outperformed so I have shown results related to this encoding of text.

Singular verbs clearly depict individuals while plural verbs depict groups and others, so their counts are included as features along with other manually extracted features listed below:
\begin{enumerate}
    \item Counts of words with positive sentiment, negative sentiment and neutral sentiment in English
    \vspace{-0.5em}
    \item Frequency of auxiliary verbs like "is", "was", "are", "were" 
    \vspace{-0.5em}
    \item Subjectivity score of the tweet (calculated using predefined libraries)
    \vspace{-0.5em}
    \item Frequency of difficult and easy words\cite{8929231}
    \vspace{-0.5em}
    \item Number of question marks, exclamations and full-stops in the tweet
    \vspace{-0.5em}
    \item Frequency of words like "they", "he", "she", "we"
    \vspace{-0.5em}
\end{enumerate}

\noindent My model for the sub-task is a neural-network based model. For the task, first, I have merged the manually-extracted features with the feature vector obtained after converting the processed tweet to one-hot encoding. The output is processed through an embedding layer which transformed the tweet into a 128 length vector. The embedding layer maps the unique indices in the one-hot vector to the embedding vector space. I pass the embeddings through a Bidirectional LSTM layer containing 256 units. This is followed by another bidirectional LSTM layer containing 512 units with its dropout and regular dropout set to 0.45 and activation being a sigmoid activation. This is followed by a Bidirectional LSTM layer with 128 units for better learning followed by a Dense Layer. The Dense layer consists of 3 units for three classes representing the classes Individual, Group and Others. The softmax layer gives a probability prediction percentage against  each of the classes and thus finally gives the final results by assigning the label with maximum probability as output. The model is compiled using the Nadam optimization algorithm with a learning rate of 0.001. Categorical crossentropy is used as the loss function.

\begin{figure}[H]
    \centering
    \includegraphics[width=0.7\textwidth]{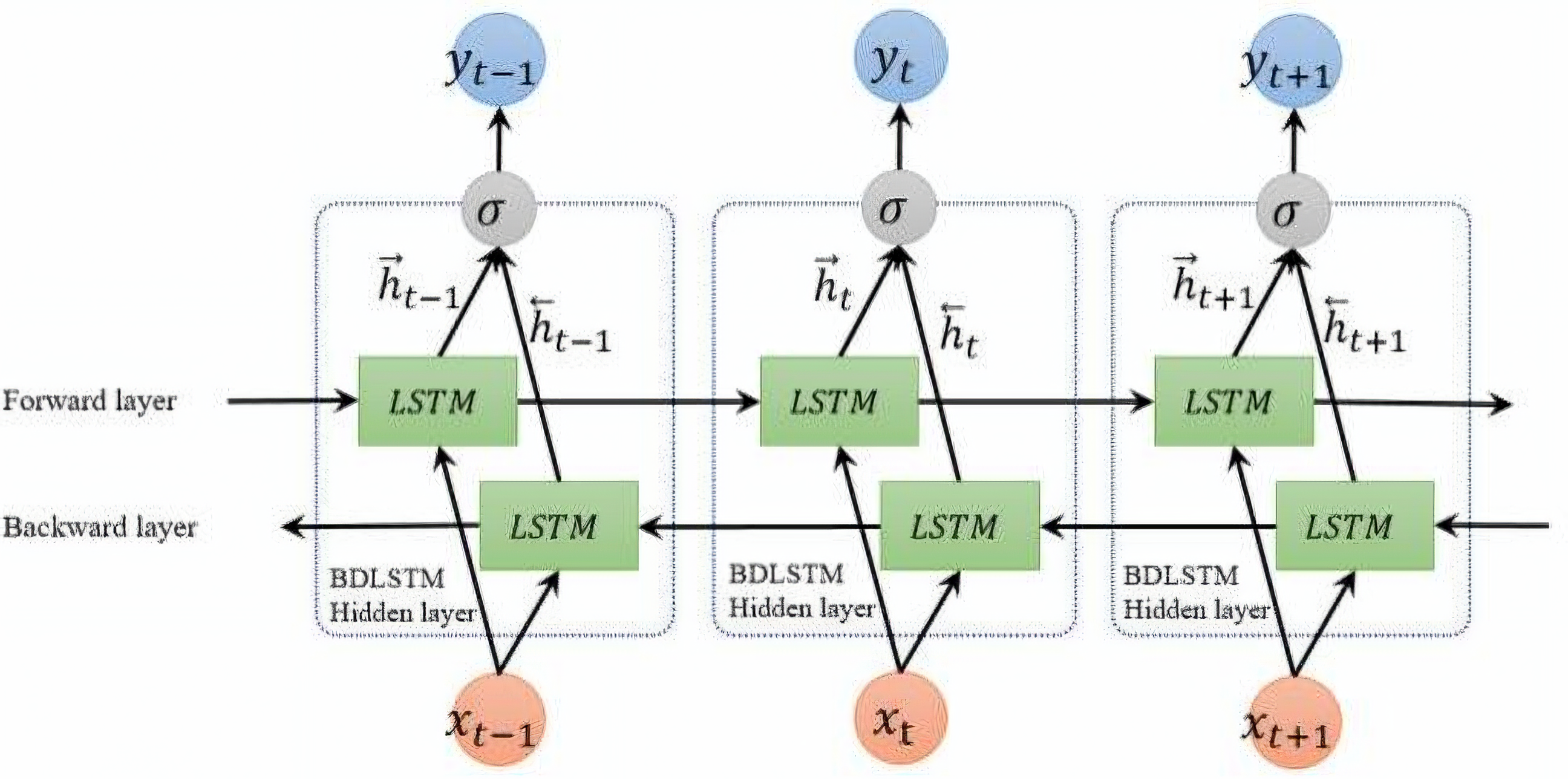}
    \caption{BiLSTM Unit}
    \label{fig:bilstm}
\end{figure}

The architecture is depicted in Table \ref{tab:working}.
\vspace{-1em}
\begin{center}
\begin{table}[H]
\begin{tabular}{lll}
\hspace{3.5cm}\_\_\_\_\_\_\_\_\_\_\_\_\_\_\_\_\_\_\_\_\_\_\_\_\_\_\_\_\_\_\_\_\_\_\_\_\_\_\_\_\_\_\_\_\_\_\_\_\_\_\_\_\_\_\_\_\_\_\_\_\_\_\_\_\_ &           \\
\hspace{4cm}Layer (type)     \hspace{0.7cm}         Output Shape        \hspace{0.8cm}      Param \#                                                                    &           \\
\hspace{3.5cm}=========================================                                    &           \\
\hspace{3.5cm}embedding\_1 (Embedding)      (None, 100, 128)          6400000                                                                    &           \\
\hspace{3.5cm}\_\_\_\_\_\_\_\_\_\_\_\_\_\_\_\_\_\_\_\_\_\_\_\_\_\_\_\_\_\_\_\_\_\_\_\_\_\_\_\_\_\_\_\_\_\_\_\_\_\_\_\_\_\_\_\_\_\_\_\_\_\_\_\_\_ &           \\
\hspace{3.5cm}bidirectional\_1 (Bidirection) (None, 100, 256)          263168                                                                     &           \\
\hspace{3.5cm}\_\_\_\_\_\_\_\_\_\_\_\_\_\_\_\_\_\_\_\_\_\_\_\_\_\_\_\_\_\_\_\_\_\_\_\_\_\_\_\_\_\_\_\_\_\_\_\_\_\_\_\_\_\_\_\_\_\_\_\_\_\_\_\_\_ &           \\
\hspace{3.5cm}bidirectional\_2 (Bidirection) (None, 100, 512)          1050624                                                                    &           \\
\hspace{3.5cm}\_\_\_\_\_\_\_\_\_\_\_\_\_\_\_\_\_\_\_\_\_\_\_\_\_\_\_\_\_\_\_\_\_\_\_\_\_\_\_\_\_\_\_\_\_\_\_\_\_\_\_\_\_\_\_\_\_\_\_\_\_\_\_\_\_ &           \\
\hspace{3.5cm}dropout\_1\hspace{.6cm} (Dropout)          (None, 100, 512)\hspace{0.8cm}          0                                                                          &           \\
\hspace{3.5cm}\_\_\_\_\_\_\_\_\_\_\_\_\_\_\_\_\_\_\_\_\_\_\_\_\_\_\_\_\_\_\_\_\_\_\_\_\_\_\_\_\_\_\_\_\_\_\_\_\_\_\_\_\_\_\_\_\_\_\_\_\_\_\_\_\_ &           \\
\hspace{3.5cm}bidirectional\_3 \hspace{.4cm}(Bidirection) (None, 256)\hspace{0.4cm}               656384                                                                     &           \\
\hspace{3.5cm}\_\_\_\_\_\_\_\_\_\_\_\_\_\_\_\_\_\_\_\_\_\_\_\_\_\_\_\_\_\_\_\_\_\_\_\_\_\_\_\_\_\_\_\_\_\_\_\_\_\_\_\_\_\_\_\_\_\_\_\_\_\_\_\_\_ &           \\
\hspace{3.5cm}dense\_1 (Dense)\hspace{1.5cm}  (None, 3)\hspace{2cm}                 514                                                                        &           \\
\hspace{3.5cm}=========================================                   &           \\
\hspace{3.5cm}Total params    \hspace{4.7cm}                                                    8,370,690                                                          &  \\
\hspace{3.5cm}Trainable params          \hspace{4cm}                                            8,370,690        &  \\
\hspace{3.5cm}Non-trainable params           \hspace{4cm}                                                       0  &          \\
\hspace{3.5cm}\_\_\_\_\_\_\_\_\_\_\_\_\_\_\_\_\_\_\_\_\_\_\_\_\_\_\_\_\_\_\_\_\_\_\_\_\_\_\_\_\_\_\_\_\_\_\_\_\_\_\_\_\_\_\_\_\_\_\_\_\_\_\_\_\_ &          
\end{tabular}
\caption{Neural Network Model Architecture}
\label{tab:working}
\end{table}
\end{center}
\vspace{-2em}
For the neural models in the language context, most popular are LSTMs (Long short term memory) which are a type of RNN (Recurrent neural network), which preserve the long term dependency of text. I use a Bidirectional-LSTM based approach to capture information from both the past and future context thus enhancing the context grasping capacity of the architecture. 
Figure \ref{fig:bilstm} shows working of a BiLSTM unit.
\vspace{-1em}

The dataset is skewed in nature. If trained on the entire dataset without any validation, the model tends to completely overfit to the class with higher frequency as it leads to a higher training accuracy score. To overcome this problem, I have taken some steps. Firstly, the training data has already been splitted into two parts - one for training and one for validation thus, solving overfitting to some extent. The training is stopped when two consecutive epochs increased the measured loss function value and decrease in Validation accuracy for the validation set. Secondly, class weights have been assigned to the different classes present in the data which are chosen to be proportional to the inverse of the respective frequencies of the classes. Hypothetically, the model then gives equal weight to the skewed classes and this penalizes tendencies to overfit to the data.

\vspace{-0.5em}
\section{Results}
\label{sec:results}
\vspace{-0.5em}
I participated in subtask C of OffenseEval 2020 which is task 12 of SemEval 2020 and my system worked quite well. I have included the automatically generated macro averaged evaluation metrics along with the detailed metrics calculated using the gold labels. The results are depicted in Tables \ref{tab:with-hastags}-\ref{tab:results-task2}.

\begin{figure}[H]
    \centering
    \includegraphics[width=0.8\textwidth]{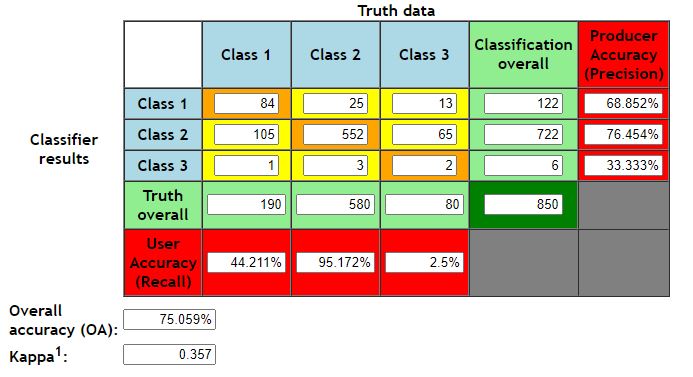}
    \caption{Confusion Matrix and Kappa value}
    \label{fig:cfm}
\end{figure}

\begin{table}[H]
\center
\begin{tabular}{|c|c|c|}
\hline
\bf System & \bf Train (\%) & \bf Validation (\%) \\ 
\hline
Without & 91.37 & 86.23 \\
\hline
With & 93.7 & 89.75 \\
\hline
\end{tabular}
\caption{Comparison of development phase accuracies with and without pre-processing operations}
\label{tab:with-hastags}
\end{table}
\vspace{-1em}
\centering \textbf{Task-C}
\begin{table}[H]
\center
\begin{tabular}{|l|l|l|l|l|}
\hline
\bf Class & \bf Precision & \bf Recall & \bf F1-score & \bf support \\ \hline
GRP   & 0.69      & 0.44   & 0.54     & 190     \\ \hline
IND   & 0.76      & 0.95   & 0.85     & 580     \\ \hline
OTH   & 0.33      & 0.03   & 0.05     & 80      \\ \hline
\textbf{macro avg}   & \textbf{0.60}      & \textbf{0.47}   & \textbf{0.4776}     & \textbf{850}     \\ \hline
\end{tabular}
\caption{Result Metrics calculated using Gold labels dataset}
\label{tab:results-task2}
\end{table}
\begin{flushleft}
\vspace{-1em}
The confusion matrix is shown in Fig. \ref{fig:cfm}.
\end{flushleft}
\vspace{-1em}

\begin{flushleft}
The skewness in the data for the OTH class has led to systematic ignorance of the system to learning features related to the class thus resulting in worse results compared to other class labels of Offensiveness.
\section{Conclusion}
\label{sec:conclusion}

In this paper, I have presented a model which performs satisfactorily in the given task. The model is based on a many2one sequence learning based architecture. There is scope for improvement by including more manually extracted features (like those removed in the pre-processing step) to increase the performance. I could use only 25\% of the whole dataset due to lack of computational resources. Data required for a Deep Learning model is quite high. Using the whole database would surely give excellent results.  Removing the data constraints might lead to better accuracies and f1 metrics. Use of regularizers can also lead to proper generalization of model, henceforth increasing the metrics.\nocite{*}
\end{flushleft}

\bibliography{ref}
\bibliographystyle{coling}

\end{document}